\documentclass[runningheads]{llncs}
\usepackage[utf8]{inputenc}
\usepackage[numbers, sort, comma, square]{natbib}
\usepackage{changepage}
\usepackage{graphicx}
\usepackage{placeins}
\usepackage{tabularx}
\usepackage{threeparttable}
\usepackage{lscape}
\usepackage{hyperref}
\usepackage{tablefootnote}
\usepackage{hyperref}
\usepackage{array}
\usepackage{booktabs}
\usepackage[normalem]{ulem}
\usepackage{caption} 
\usepackage{tabularx}

\newcolumntype{R}[1]{>{\raggedleft\let\newline\\\arraybackslash\hspace{0pt}}m{#1}}
\begin{document}
\title{Document Embedding for Scientific Articles:\newline Efficacy of Word Embeddings vs TFIDF}
\titlerunning{Document embedding for Scientific Articles}
\author{H.J. Meijer\inst{1,2}\orcidID{0000-0003-2901-8119} \and
J. Truong\inst{2}
\and
R. Karimi\inst{2}\orcidID{0000-0002-2534-1907} }
\authorrunning{H.J. Meijer et al.}
\institute{University of Amsterdam, Science park 904, 1012~WX Amsterdam, The Netherlands \and
Elsevier BV, Radarweg 29, 1043~NX Amsterdam, The Netherlands
\email{meijerarjan@live.nl, j.truong@elsevier.com, reza.karimi@gmail.com}}
\maketitle
\begin{abstract} Over the last few years, neural network derived word embeddings became popular in the natural language processing literature. Studies conducted have mostly focused on the quality and application of word embeddings trained on public available corpuses such as Wikipedia or other news and social media sources. However, these studies are limited to generic text and thus lack technical and scientific nuances such as domain specific vocabulary, abbreviations, or scientific formulas which are commonly used in academic context. This research focuses on the performance of word embeddings applied to a large scale academic corpus. More specifically, we compare quality and efficiency of trained word embeddings to TFIDF representations in modeling content of scientific articles. We use a word2vec skip-gram model trained on titles and abstracts of about 70 million scientific articles. Furthermore, we have developed a benchmark to evaluate content models in a scientific context. The benchmark is based on a categorization task that matches articles to journals for about 1.3 million articles published in 2017. Our results show that content models based on word embeddings are better for titles (short text) while TFIDF works better for abstracts (longer text). However, the slight improvement of TFIDF for larger text comes at the expense of 3.7 times more memory requirement as well as up to 184 times higher computation times which may make it inefficient for online applications. In addition, we have created a 2-dimensional visualization of the journals modeled via embeddings to qualitatively inspect embedding model. This graph shows useful insights and can be used to find competitive journals or gaps to propose new journals.
\end{abstract}
\section{Introduction}
Neural network derived word embeddings are dense numerical representations of words that are able to capture semantic and syntactic information\cite{mikolov2013distributed}. Word embedding models are calculated by capturing word relatedness\cite{hou2018enhanced} in a corpus as derived from contextual co-occurrences. They have proven to be a powerful tool and have attracted the attention of many researchers over the last few years. The usage of word embeddings has improved various natural language processing (NLP) areas including named entity recognition\cite{do2018evaluating}, part-of-speech tagging\cite{santos2014learning}, and semantic role labelling\cite{he2018jointly, luong2013better}. Word embeddings have also given promising results on machine translation\cite{gouws2015bilbowa}, search\cite{ganguly2015word} and recommendation\cite{musto2016learning, ozsoy2016word}. 
\\Similarly, there are many potential applications of the embeddings in the academic domain such as improving search engines, enhancing NLP tasks for academic texts, or journal recommendations for manuscripts. Published studies have mostly focused on generic text like Wikipedia\cite{levy2014neural,park2018conceptvector}, or informal text like reviews\cite{dos2014deep, lauren2018generating} and tweets\cite{masino2018detecting,yang2018using}. We aim to validate word embedding models for academic texts containing technical, scientific or domain specific nuances such as exact definitions, abbreviations, or chemical/mathematical formulas. We will evaluate the embeddings by matching articles to their journals. To quantify the match, we use the ranks derived by sorting similarity of embeddings between each article and all journals. Furthermore, we plot the journal embeddings as a 2-dimensional representation of journal relatedness. Our 2-dimensional plot of embeddings visualizes relatedness in a scatter plot\cite{dai2015document, hinton2003stochastic}.

\section{Data and environment}
In this study, we compare content models based on TFIDF, embeddings, and various combinations of both. This section describes the training environment and parameters as well as other model specifications to create our content models. 

\subsection{Dataset:}
Previous studies have highlighted the benefits of learning embeddings in a similar context as they are later used in\cite{lai2016generate, Truong2017Thesis}. Thus, we trained our models on title and abstracts of approximately 70 million scientific articles from 30 thousand distinct sources such as journals and conferences. All articles are derived from Scopus abstract and citation database \cite{url_scopus}. After tokenizing, removal of stopwords and stemming the dataset contains a total of ca.~5.6 billion tokens (ca.~0.64 million unique tokens). The word occurrences in this training set follow a Pareto-like distribution as described by Wiegand et al\cite{wiegand2018word}. This distribution indicates that our original data has similar properties as standard English texts.

\subsection{TFIDF}
We used 3 TFIDF alternatives all created by the TFIDF and the hasher from the pySpark mllib\cite{url_spark}. We controlled TFIDF alternatives in two ways, (a) adjusting vocabulary size and (b) adjusting the number of hash buckets. We label the TFIDF alternatives as follows:``vocabulary-size / number-of-hash-buckets''. Thus, we label the TFIDF configuration that has a vocabulary size of 10,000 and 10,000 hash buckets as TFIDF 10K/10K. To select the TFIDF sets, we measured memory footprint of multiple TFIDF configurations vs our accuracy metric (see section 3 for detailed definition). As seen in Table~\ref{table:tfidfperformance}, the performance on both title and abstract stagnates; the same is true for the memory usage. Given these results, we selected the 10K/10K, 10K/5K and 5K/5K configurations for our research as reasonable compromises between memory footprint and accuracy. We also do not expect significantly better performance for higher vocabulary sizes such as 20K.
\newcolumntype{L}[1]{>{\raggedright\arraybackslash}p{#1}}
\newcolumntype{C}[1]{>{\centering\arraybackslash}p{#1}}
\begin{table}
\caption{TFIDF accuracy and memory usage vs variable hash-bucket and vocabulary size} 
\centering
\begin{tabular}{L{8cm}|R{1.5cm}||C{1.2cm}|C{1.2cm}}
& \textbf{Memory} &  \multicolumn{2}{c}{\textbf{Median Rank}} \\
\textbf{Vocabulary Size and Number of Hash Buckets} &\multicolumn{1}{c||}{(GB)} & Title& Abstract\\
\bottomrule
1k (1k/1k)& 5.13 & 183 & 44\\
4k (4k/4k)& 9.29 & 59 & 20\\
7k (7k/7k)& 10.85 & 42 & 16\\
10k (10k/10k)& 11.61 & 35 & 14\\
\end{tabular}
\label{table:tfidfperformance}
\end{table}
\subsection{Embeddings}\label{embeddings}
Our word embeddings are obtained using a spark implementation\cite{url_spark} of the word2vec skip-gram model with hierarchical softmax as introduced by Mikolov et al\cite{mikolov2013efficient}. In this shallow neural network architecture, the word representations are the weights learned during a simple prediction task. To be precise, given a word the training objective is to maximize the mean log-likelihood of its context. We have optimized model parameters by means of a word similarity task using external evaluation sets\cite{Finkelstein2001PlacingSI, bruni2014multimodal, Luong2013BetterWR} and consequently used the best performing model (see \ref{modelopt}) as reference embedding model in this entire article (referred to as \textit{embedding}). Additionally, we created 4 variants of TFIDF combined with embedding. All embedding models are listed below: 
\newcolumntype{L}[1]{>{\raggedright\arraybackslash}p{#1}}
\newcolumntype{C}[1]{>{\centering\arraybackslash}p{#1}}
\begin{table}
\centering
\begin{tabular}{L{3cm} L{9cm}}
\textit{- embedding:} & unweighted mean embedding of all tokens \\
\textit{- TFIDF\_embedding:} & tfidf-weighted mean embedding of all tokens \\
\textit{- 10K\_embedding:} & tfidf-weighted mean of the top 10,000 most occurring tokens\\
\textit{- 5K\_embedding:} & tfidf-weighted mean embedding of the top 5,000 most occurring tokens\\
\textit{-1K\_6K\_embedding}: & tfidf-weighted mean embedding of the top 6,000 most occurring tokens excluding the 1,000 most occurring tokens.
\end{tabular}
\label{table:modelvariants}
\end{table}

\section{Methodology}
To measure the quality of embeddings, we calculate a ranking between each article and its corresponding journal. This ranking, calculated by comparing embedding of the article with the embedding of all journals, will resemble the performance of embeddings in a categorization task. Articles in 2017 are split into 80\%-20\% training and test sets. Within training set, we average embeddings per journal and define it as the embedding per journal. This study is limited to journals with at least 150 publications in 2017 and those who had papers in both test and training sets (roughly 3700 journals and 1.3 millions articles). We calculate the similarity of embeddings between each article in the test set and all journals in the training set. We order similarity scores such that rank one corresponds to the journal with the most similar embedding. We record the rank of the source journal of each article for evaluations. We do this for both title and abstract separately. We calculate the performance per set, therefore we combine the ranking results of all articles for a set into one score. We use the median and average for that: the average rank takes the total average of all ranks, while the median is the point at which 50\% of all the ranks are higher and 50\% of all the ranks are lower. We keep track of the following results when ranking: source journal rank, score as well as name of the best matching journal for both abstract and title. We furthermore monitor the memory usage and computation time. To plot the journal embeddings, we use PCA (Principal Component Analysis)-based tSNE. tSNE (t-Stochastic Neighbor Embedding) is a vector reduction method introduced by Maaten et al\cite{maaten2008visualizing}.
\section{Results}
In this section, the results of our research are presented; the detailed discussion on the meaning and implications of these results are presented in section 5, Discussion.
\subsection{Model Optimization} \label{modelopt}
During optimization, we tested the effect of several learning parameters on training time and quality using three benchmark sets for evaluating word relatedness, i.e.~the WordSimilarity-353 \cite{Finkelstein2001PlacingSI}, MEN \cite{bruni2014multimodal} and the Rare Word \cite{luong2013better} test collection that contain generic and rare English word pairs along with human-assigned similarity judgments. Only few parameters, i.e. number of iterations, vector size and the minimum word count for a token to be included in the vocabulary had significant effect on the quality. The learning rate was 0.025 and the minimum word count was 25. Our scores were close to external benchmarks from above studies. We manually investigated the differences and they were mostly due to word choice differences between academic context vs non-academic. Indeed, the biggest difference was between television and show pairs (because in academic context show would rarely relate to television). Table \ref{embeddings} contains the average scores and training times when tuning the parameters while fixing the remaining one. Our final and reference model is based on 300-dimensional vectors, a context window of 5 tokens, 1 iteration and 160 partitions.  
\begin{table}
\caption{average accuracy scores and computation times during training.}
\centering
\begin{threeparttable}
\begin{tabular}{L{4cm}|C{1.5cm}|C{2.5cm}|C{2.5cm}}
& \textbf{value} & \textbf{average score} &  \textbf{training time}\\
\bottomrule
vector size & 100 & 0.447 & 3.2 h\\
 & 200 & 0.46 & -\tnote{a} \\
 & 300 & 0.51 & -\tnote{a} \\
\hline
no. of iteration & 1 & 0.446 & 2.94 h\\
 &  3 & 0.457 & 4.48 h \\
 &  6 & 0.46 & 7.1 h \\
\hline
min. word count & 15 & 0.467 & -\tnote{b} \\
 &  25 & 0.473 & -\tnote{b} \\
 &  50 & 0.447 & -\tnote{b} \\
\end{tabular}
\begin{tablenotes}
\item[a] ran in different cluster due to memory issues.
\item[b] not significant.
\end{tablenotes}
\end{threeparttable}
\label{table:embeddings}
\end{table}
\subsection{Ranking}
\begin{figure}
\includegraphics[width=4.5in]{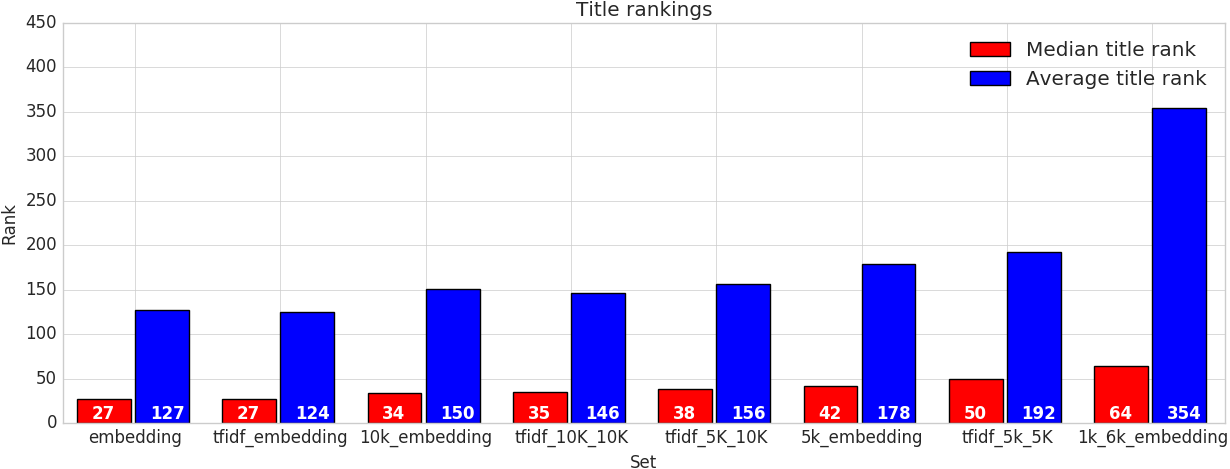}
\caption{Median and average title rankings}\label{figure:titleRanks}
\includegraphics[width=4.5in]{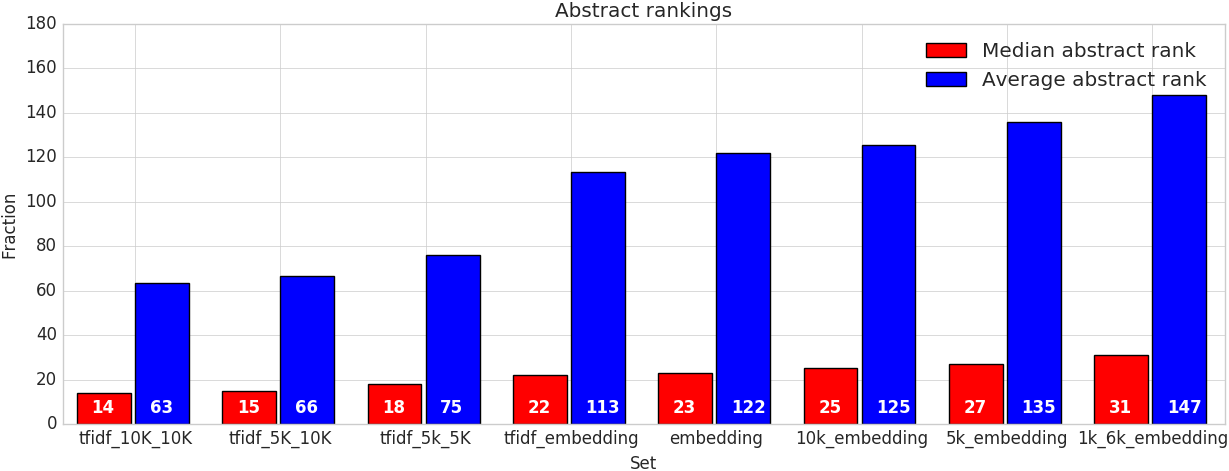}
\caption{Median and average abstract rankings}\label{figure:abstractRanks}
\end{figure}

Figures~\ref{figure:titleRanks} and \ref{figure:abstractRanks} show the result of the categorization task via ranking measures for titles and abstracts. The rank indicates the position of the correct journal in the sorted list of all journals. These graphs show both the average and the median ranks, based on the cosine-similarity between the article and journal embeddings. These embedding vectors, whether calculated by word2vec, TFIDF or their combinations can be considered as the feature vectors used elsewhere for machine-learning tasks.
\begin{figure}
\includegraphics[width=4.5in]{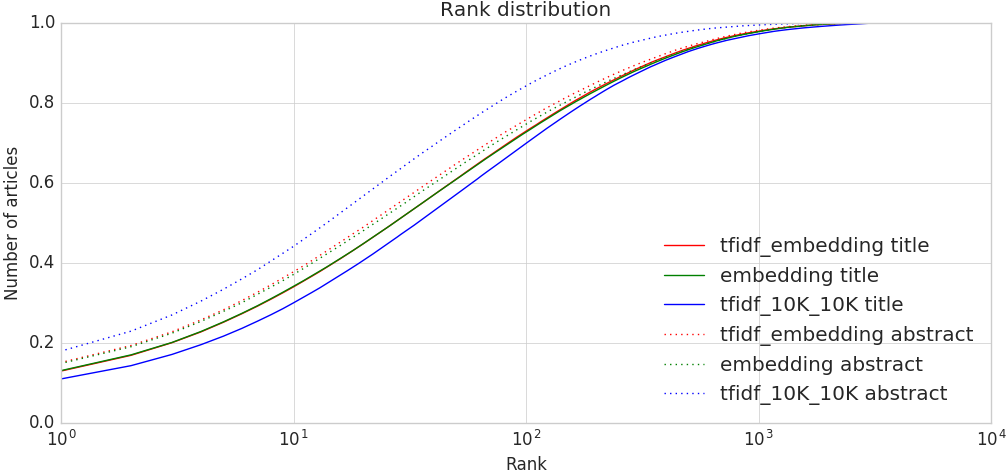}
\caption{Rank distribution for the title and the abstract: Y-axis shows the fraction of articles.}\label{figure:distributions}
\end{figure}
\subsection{Rank Distribution}
Figure~\ref{figure:distributions} shows the distributions of the ranks for our default embeddings, TFIDF weighted embedding and the 10K/10K TFIDF. The figure plots the cumulative percentage of articles as a function of rank. The plot gives a detailed view of the ranks  presented in Figures~\ref{figure:titleRanks} and~\ref{figure:abstractRanks}.\\
\subsection{Memory Usage and Computation Time}
Table~\ref{table:memoryUsage} shows the total memory usage of each test set in gigabytes. Moreover, it provides the absolute hit percentage of the title and the abstracts, i.e. the percentage of articles that have their source journal as the first result in the ranking. The table furthermore lists the median rank and the median abstract rank, as visualized in Figures \ref{figure:titleRanks} and~\ref{figure:abstractRanks}. Thus, this table gives an overview of the memory usage of the sets, combined with their accuracy on the ranking task.\\
We furthermore investigated compute efficiency for different content models. To simulate what can happen during an online application, we selected 1000 random articles and then measured time needed for dot products between all pairs. Time recorded excluded input/output time and all calculations started from cached data with optimized numpy matrix/vector calculations. Table~\ref{table:computationTimes} shows computation time in seconds as well as ratios. Generally dot products can be faster for dense vectors as opposed to sparse vectors. Generally TFIDF vectors are stored as sparse vectors while embeddings are dense vectors. Hence, we also created a dense vector version of TFIDF sets to isolate the effect of sparse vs dense representation.
\begin{table}
\caption{memory usage and performance for various content models}
\centering
\begin{tabular}{L{4cm}|C{1.5cm}||C{1.5cm}|C{1.5cm}||C{1.5cm}|C{1.5cm}}
& \textbf{Memory} &  \multicolumn{2}{c||}{\textbf{Absolute Hit}}  & \multicolumn{2}{c}{\textbf{Median Rank}} \\
&\multicolumn{1}{c||}{(GB)}&Title & Abstract& Title& Abstract\\
\bottomrule
tfidf 5k 5K & $9.82$ & 5.42\% & 10.18\% & 50 & 27\\
tfidf 5K 10K & $11.47$ & 6.49\% & 11.08\% & 38 & 15\\
\textbf{tfidf 10K 10K} & 11.61& 6.79\% & \textbf{11.32\%}& 35 & \textbf{14}\\
\textbf{embedding} & $3.13$ & \textbf{7.92}\% & 9.24\% & \textbf{27} & 23\\
5k embedding & $3.13$ & 6.34\% & 8.36\% & 42 & 27\\
10k embedding & $3.13$ & 7.03\% & 8.76\% & 34 & 25\\
\textbf{tfidf embedding} & 3.13 & 7.89\% & 9.33\% & \textbf{27} & 22\\
1k 6k embedding & $\textbf{3.06}$ & 5.16\% & 7.86\% & 64 &31 \\
\end{tabular}
\label{table:memoryUsage}
\end{table}
\begin{table}
\caption{comparing computation time between embeddings and TFIDF models}
\centering
\begin{tabular}{L{4cm}|R{1.5cm}|R{1.5cm}||R{1.5cm}|R{1.5cm}}
  &  \multicolumn{2}{c|}{\textbf{Seconds}} & \multicolumn{2}{c}{\textbf{Ratio vs Embedding}}  \\
 & Title & Abstract & Title & Abstract \\
\bottomrule
TFIDF (sparse vector) &   154.95 & 169.89  & 231.25 & 184.36\\
TFIDF (dense vector) & 35.67 & 35.18  &  53.23 & 39.59 \\
\textbf{Embedding} &  \textbf{0.67} & \textbf{0.89} & \textbf{1} & \textbf{1}\\
\end{tabular}
\label{table:computationTimes}
\end{table}
\subsection{Journal Plot}
\begin{landscape}
\begin{figure}
\begin{center}
\includegraphics[height=4.34in]{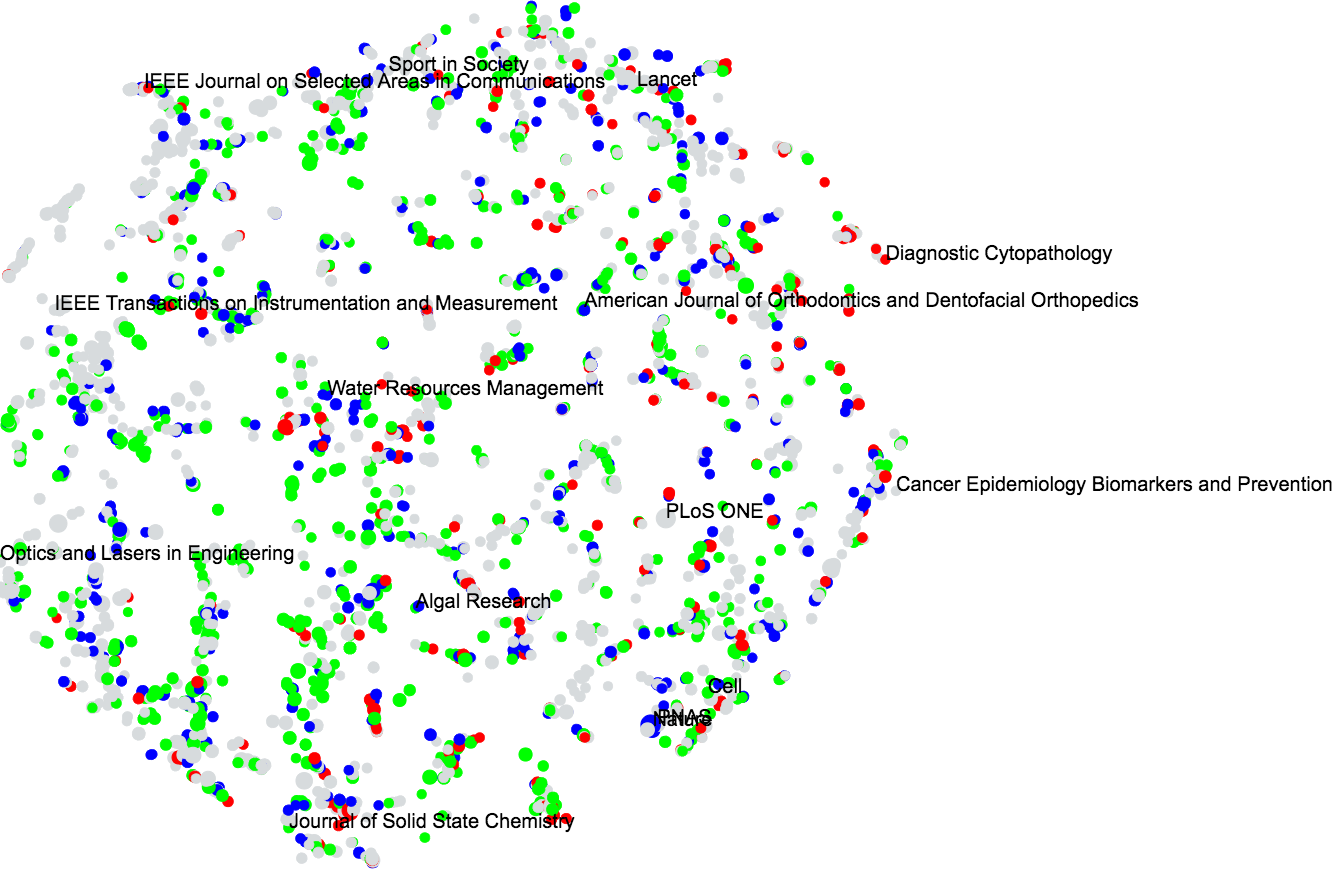}
\end{center}
\caption{Journal plot of abstract embeddings after tSNE transformation. Red, green, blue, and gray represent respectively Wiley, Elsevier, Springer-Nature, and other/unknown publishers. }\label{figure:abstractPlotNormal}
\end{figure}
\end{landscape}

Figure~\ref{figure:abstractPlotNormal} shows the 2-dimensional visualization of the (default) journal embeddings based on the abstracts. This plot is color coded to visualize publishers. Some journal names have been added to the plot to indicate research areas.
\section{Discussion}
\subsection{Results Analysis}
\subsubsection{Highest Accuracy}
The data, as presented in Figures~\ref{figure:titleRanks} and~\ref{figure:abstractRanks} shows that the 10k/10k set performs better than all other TFIDF sets, although the difference with the 5k/10k is low (a median rank difference of 1 on the abstracts and 3 on the titles). For the embeddings, the TFIDF weighted embedding outperforms other embedding models by a narrow margin: 1 median rank higher on the abstracts, and equal on the titles. 

\subsubsection{Dataset and Model Optimization}
The determinants for choosing the final model parameters were constrained by their computational costs. Hence, even if increasing the number of iterations could have led to better performing word embeddings we chose 1 training iteration due to the increased training time. Similarly, we decided to stem tokens prior to training in order to decrease the vocabulary size. This might have led to a loss of syntactical information or caused ambiguous tokens.

\subsubsection{TFIDF}
The TFIDF feature vectors outperform the embeddings on abstracts, while the embeddings outperform the TFIDF on titles. The main difference between abstract and title is the number of tokens. Hence, embeddings which enhance tokens by their semantic meaning outperform TFIDF on the title. On the other hand, the TFIDF model outperforms on the abstract likely due to additional specification by additional tokens. In other words, longer text provides a better context and hence requires a less accurate semantic model for individual tokens. 
\\Furthermore, none of our various vocabulary size cut-offs improved TFIDF ranks and indeed increasing the vocabulary size monotonically increased the performance of the TFIDF. In other words, we could not find a cut-off strategy to reduce noise and enhance TFIDF results. Although, it could still be possible that at even higher vocabulary sizes the cut-off would result in a sharper signal. However, since we noticed performance stagnation we did not investigate larger vocabulary sizes beyond 10k (presented in Table~\ref{table:tfidfperformance}).

\subsubsection{Combination of TFIDF and embeddings}
The limited TFIDF embeddings all fall short of full TFIDF embedding. We did not find a vocabulary size cut-off strategy to increase accuracy by reducing noise from rare or highly frequent words or their combinations. In other words, it is best not to miss any word. This is in line with what we found with the TFIDF results: larger vocabulary sizes enhances models. 
\\\textit{Rank distribution}; Although the limited TFIDF embeddings underperform, we found that their rank distribution is different from the other embeddings. The rank distribution of the limited TFIDF embeddings shows the following pattern: a high/average performance on the top-rankings, a below average performance on the middle rankings and an increased ratio of articles with worsened higher ranks. 
The rank distribution seems to indicate that the cut-offs marginalize ranks. The cut-off moved the ``middle-ranked'' articles to either the higher end or the lower end with a net effect to deteriorate median ranks. However, the articles that matched with limited TFIDF embedding had higher accuracy scores. The reduction in vocabulary size did not reduce the storage size for the embeddings, except for the 1K-6K case. This indicated that only the 1K-6K cuts removed all tokens from some abstract and titles resulting in null records and hence lower memory.
\subsubsection{TFIDF \& embeddings}
Our hypothesis on the difference between the TFIDF and the standard embedding is as follows:\\
The embeddings seem to outperform the TFIDF feature vectors in situations where there is little information available (titles). This indicates that the embeddings store some word meaning that enables them to perform relatively well on the titles. The abstracts, on the other hand, contain much more information. Our data seems to indicate that the amount of information available in the abstracts enables the TFIDF to cope with the lack of an explicit semantic model. 
If this is the case, we could expect that there would be little performance increase on the title when we compare the embeddings to the weighted TFIDF embeddings, because the TFIDF lacks the information to perform well. This can be seen in our data, only the average rank increased by 3, indicating that there is a difference between the two embeddings, but not a significant one. We would also expect on the abstract an increase in performance since the TFIDF has more information in this context. We would expect that the weighting applied by the TFIDF model, an importance weighting, will improves the performance of the embedding. Our data shows a minor improvement in performance: 1 median rank and 10 average ranks. While these improvements cannot be seen as significant, our data at least indicates that weighting the embeddings with TFIDF values has a positive effect on the embeddings.
\subsubsection{Memory usage \& Calculation time}
TFIDF outperforms the embeddings on the abstracts, but requires more memory. Embedding uses 3.13 GB RAM, while the top performing TFIDF, 10K/10K, uses 11.61 GB (3.7 times more RAM footprint). This indicates that the embeddings are able to store the relatedness information more densely than the TFIDF. The embeddings furthermore need less calculation time for online calculations as shown in Table~\ref{table:computationTimes}. In average, embeddings are 200 times faster than sparse TFIDF. When the vectors are transformed to dense vectors this is reduced to 46 times. The difference between the sparse and dense vectors is due to dense vectors being processed more efficiently by low level vector operations. The difference between the embedding and TFIDF dense vectors is mainly due to the vector size. Embeddings use a 300 dimensional vector, while TFIDF uses a 10000 dimensional vector. Hence a time ratio of 33 is just normal and indeed close to measured values of 40 and 53 in Table ~\ref{table:computationTimes}. Note that even though dense representation is roughly 4-5 times faster, it requires 33 times more RAM which can be prohibitive.
\subsection{Improvements}
This research demonstrates that even though the embeddings can capture and preserve relatedness, TFIDF is able to outperform the embeddings on the abstracts. We used basic word2vec but earlier research already shows additional improvement potential for word2vec. Dai et al\cite{dai2015document} showed that using paragraph vectors improves the accuracy of word embeddings with 4.4\% on triplet creation with the Wikipedia corpus and a 3.9\% improvement on the same task based on the arXiv articles. Furthermore, Le et al\cite{le2014distributed} show that the usage of paragraph vectors decrease the error rate (positive/negative) with 7.7\% compared to averaging the word embeddings on categorizing text as either positive or negative. While the improvement looks promising, we have to keep in mind that our task differs from earlier research. We do not categorize on two categories but about 3700 journals. Since the classification task is fundamentally the same, still we would expect an improvement by using paragraph vectors. However, the larger scale here complicates the task due to the ``grey areas'' between categories. These are the areas in which the classification algorithm is ``in doubt'' and could reasonably assign the article to both journals. There are many similar journals and hence we cannot expect a rank 1 for most of articles. Indeed our classes here are not exactly mutually exclusive. Indeed in general, the number of these grey areas increase with increased target class size.\\ Pennington et al\cite{pennington2014glove} showed that the GloVe model outperforms the continuous-bag-of-words (CBOW) model, which is used in this research, on a word analogy task. Wang et al\cite{wang2016linked} introduced the linked document embedding method (LDE) method, which makes use of additional information about a document, such as citations. Their research specifically focused on categorizing documents, showed a 5.89\% increase of the micro-F1 score on LDE compared to CBOW, and a 9.11\% increase of the macro-F1 score. We would expect that applying this technique to our dataset would improve our scores, given earlier results on comparable tasks. Although our results seem to indicate that the embeddings work for academic texts, Schnabel et al\cite{schnabel2015evaluation} found that the quality of the embeddings are depended on the validation task. Therefore, conservatively we can only state that our research shows that embeddings work on academic texts for journal classifications.\\
Despite immense existing researches, we have not been able to find published results which are directly comparable to ours. This is due to our large target class size (3700 journals) that requires a ranking measure. Earlier research limited themselves to small number of groups such as binary classes or 3 classes \cite{shen2018baseline}. We have opted median rank as our key measure, but like existing research we have also reported absolute hit \cite{wang2016linked}. Our conclusions, were indifferent to exact metric used (median vs average rank vs absolute hit).
\section{Conclusion}
Our research, based on academic corpus, indicates that embeddings provide a better content model for shorter text such as title and fall short of TFIDF for larger texts such as abstracts. The higher accuracy of TFIDF may not be worth it, as it requires 3.7 more RAM and is roughly 200 times slower for online applications.
The performance of the embeddings have been improved by weighing them with the TFIDF values on the word level, although this improvement cannot be seen as significant on our dataset. The visualization of the journal embedding shows that similar journals are grouped together, indicating a preservation of relatedness between the journal embeddings. 
\section{Future work}
\subsubsection{Intelligent cutting}
A better way of cutting could improve the quality of the embeddings. This improvement might be achieved by cutting the center of the vector space out before normalization. All words which are generic are in the center of the spectrum, removing these words prevents the larger texts to be pulled towards the middle of the vector space, where they lose the parts of their meaning which set them apart from the other texts. We expect that this way of cutting, instead of word-occurrence cutting, can enhance embeddings especially for longer texts.
\subsubsection{TFIDFs performance point}
In our research, TFIDF performed better on the abstracts than on the titles, which we think is caused by the difference in text size.  Consequently, there could be a critical length of text where the best performing model switches from embedding to TFIDF. If this length is found, one could skip the TFIDF calculations in certain situations, and skip the embedding training in other scenario's, reducing costs.
\subsubsection{Reversed word pairs}
At this point, there are no domain-specific word pair sets available. However, as we demonstrated, we can still test the quality of word embeddings. Once one has established that the word vectors are of high quality, could one create word pairs from these embeddings? If this is the case we could create word pair sets using the embeddings and then reverse engineer domain specific word pairs for future use.
\section{Acknowledgement}
We would like to thank Bob JA Schijvenaars for his support, advice and comments during this project. 
\bibliographystyle{splncs04}
\bibliography{references}
\end{document}